\documentclass[letterpaper]{article} 
\usepackage{aaai2026}  
\usepackage{times}  
\usepackage{helvet}  
\usepackage{courier}  
\usepackage[hyphens]{url}  
\usepackage{graphicx} 
\usepackage{xcolor}  
\usepackage{amssymb} 
\usepackage{bm} 
\usepackage{booktabs} 
\usepackage{multirow} 
\usepackage{amsmath} 
\usepackage{natbib}  
\usepackage{caption} 
\usepackage{algorithm}
\usepackage{algorithmic}

\usepackage{newfloat}
\usepackage{listings}
\DeclareCaptionStyle{ruled}{labelfont=normalfont,labelsep=colon,strut=off} 
\floatstyle{ruled}
\newfloat{listing}{tb}{lst}{}
\floatname{listing}{Listing}
\title{ActiShade: Activating Overshadowed Knowledge to Guide Multi-Hop Reasoning in Large Language Models}
\author{
    Huipeng Ma\textsuperscript{\rm 1,2}\equalcontrib,
    Luan Zhang\textsuperscript{\rm 1}\equalcontrib ,
    Dandan Song\textsuperscript{\rm 1}\thanks{Corresponding author.},
    Linmei Hu\textsuperscript{\rm 1}\footnotemark[2],
    Yuhang Tian\textsuperscript{\rm 1},
    Jun Yang\textsuperscript{\rm 1},
    Changzhi Zhou\textsuperscript{\rm 1},
    Chenhao Li\textsuperscript{\rm 1},
    Yizhou Jin\textsuperscript{\rm 1},
    Xudong Li\textsuperscript{\rm 1},
    Meng Lin\textsuperscript{\rm 1},
    Mingxing Zhang\textsuperscript{\rm 3},
    Shuhao Zhang\textsuperscript{\rm 4}
}
\affiliations{
    \textsuperscript{\rm 1}Beijing Institute of Technology, China\\
    \textsuperscript{\rm 2}QiYuan Lab \\
    \textsuperscript{\rm 3}Tsinghua University, China \\ 
    \textsuperscript{\rm 4}Huazhong University of Science and Technology, China\\

    \{mahuipeng, luan\_zhang, sdd, hulinmei\}@bit.edu.cn 
}

\usepackage{bibentry}

\begin{document}

\maketitle

\begin{abstract}
In multi-hop reasoning, multi-round retrieval-augmented generation (RAG) methods typically rely on LLM-generated content as the retrieval query. 
However, these approaches are inherently vulnerable to \textit{knowledge overshadowing}—a phenomenon where critical information is overshadowed during generation.  
As a result, the LLM-generated content may be incomplete or inaccurate, leading to irrelevant retrieval and causing error accumulation during the iteration process. 
To address this challenge, we propose \textbf{ActiShade}, which detects and activates overshadowed knowledge to guide large language models (LLMs) in multi-hop reasoning.
Specifically, ActiShade iteratively detects the overshadowed keyphrase in the given query, retrieves documents relevant to both the query and the overshadowed keyphrase, and generates a new query based on the retrieved documents to guide the next-round iteration. 
By supplementing the overshadowed knowledge during the formulation of next-round queries while minimizing the introduction of irrelevant noise, ActiShade reduces the error accumulation caused by \textit{knowledge overshadowing}. 
Extensive experiments show that ActiShade outperforms existing methods across multiple datasets and LLMs. 
\end{abstract}


\section{Introduction}

Large language models (LLMs) have demonstrated remarkable performance across various of NLP tasks, such as multi-hop reasoning~\cite{DBLP:journals/corr/abs-2303-08774, meta2024introducing}. 
However, LLMs have a risk of generating factually incorrect responses, also known as hallucinations~\cite{bang-etal-2023-multitask,ji2023survey,huang2023survey}.
Retrieval-augmented generation (RAG) techniques have been widely adopted to enhance the factual correctness of LLM-generated responses by incorporating knowledge from external resources~\cite{DBLP:journals/corr/abs-2312-10997, DBLP:conf/kdd/FanDNWLYCL24}. 
\definecolor{myPurple}{HTML}{9673A6}  
\definecolor{mygreen}{HTML}{82B366}  
\definecolor{myblue}{HTML}{6C8EBF}  
\begin{figure}[h]
\centering
\includegraphics[scale=0.38]
{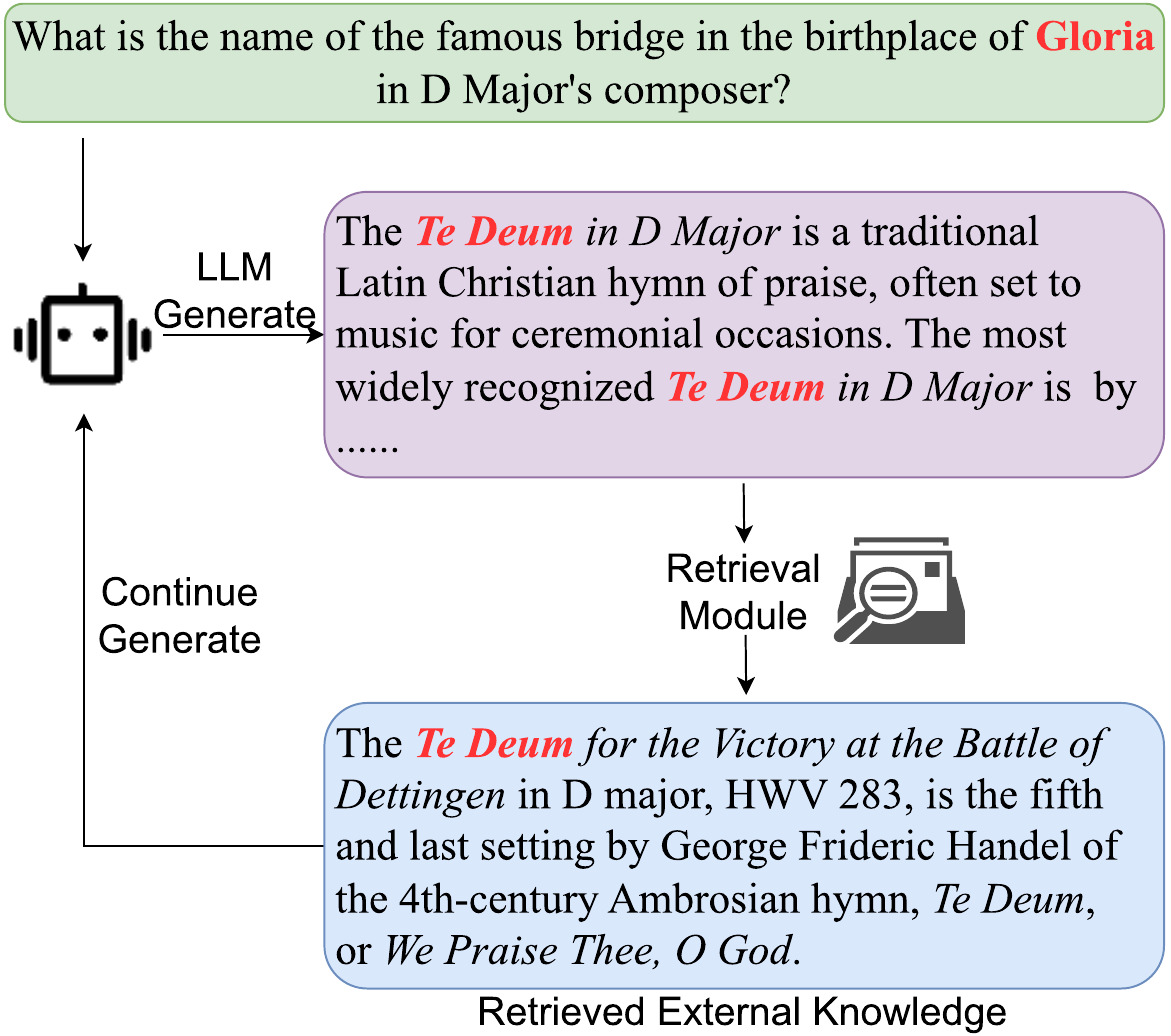}
\caption{Illustration of error accumulation caused by \textit{knowledge overshadowing}.The keyphrase Gloria in the query is overshadowed, leading the LLM to generate inaccurate content, such as Te Deum. This results in the retrieval of irrelevant documents, which in turn causes LLM to generate more inaccurate content in the next-round iteration.}  
\label{fig:ss}
\end{figure}

Early RAG methods often adopt one-round retrieval, \textit{i.e.}, use the original question as the retrieval query~\cite{DBLP:conf/icml/GuuLTPC20, DBLP:conf/icml/BorgeaudMHCRM0L22, DBLP:journals/jmlr/IzacardLLHPSDJRG23, DBLP:conf/emnlp/ZhangZRSHWLC23}. Although these methods show satisfactory performance in answering single-hop questions~\cite{DBLP:conf/acl/JoshiCWZ17, DBLP:journals/tacl/KwiatkowskiPRCP19}, they fail in answering multi-hop questions, where more knowledge is needed beyond the one-round retrieved knowledge. 

Recent research proposes multi-round retrieval, which typically relies on the LLM-generated content to guide subsequent-round retrieval. 
A possible approach uses the response generated by LLMs for retrieval and, in turn, uses the newly retrieved knowledge for generation. 
By iteratively alternating between retrieval-augmented generation and generation-augmented retrieval, the retrieval and generation are improved~\cite{iter-retgen, ircot, flare, su-etal-2024-dragin}. 
Another approach prompts LLMs to decompose the  complex question into a sequence of sub-questions, using the sub-question as the retrieval query to obtain more precise knowledge~\cite{self-ask, DBLP:conf/iclr/ZhouSHWS0SCBLC23, DBLP:conf/emnlp/CaoZSL0THL23, DBLP:conf/acl/ChuCCWZDYLQ24}. 

However, these methods suffer from \textit{knowledge overshadowing}~\cite{zhang2024knowledge} — a phenomenon where dominant conditions can overshadow others, causing the LLM to overlook essential information during generation. As illustrated in Figure~\ref{fig:ss}, given the original query, the dominant condition \textit{Te Deum in D Major} overshadows the condition \textit{Gloria in D Major}, causing the LLM to generate next-round query related to \textit{Te Deum in D Major}.
As a result, it retrieves irrelevant documents, which mislead the LLM in generating the subsequent query, ultimately leading to error accumulation over multi-round iterations. This phenomenon is particularly problematic in multi-hop reasoning, where the reasoning process relies on multiple interrelated conditions within the query. 

Motivated by this, we propose \textbf{ActiShade}, a novel
framework designed to detect and subsequently leverage overshadowed knowledge, thereby reducing error accumulation in multi-hop reasoning.  
ActiShade first detects the overshadowed knowledge within the query, then retrieves documents relevant to it, enabling LLMs to focus on critical but overlooked information during reasoning. 
Specifically, ActiShade consists of three modules. 
({\romannumeral1}) \textit{Knowledge Overshadowing Detection}: We design a new Gaussian perturbation-based method (\textbf{GaP}), to detect overshadowed keyphrases by perturbing the embeddings of candidate keyphrases with Gaussian noise and assessing the changes in the LLM's output distribution.
({\romannumeral2}) \textit{Retrieval based on Overshadowed Keyphrase}: 
We train a dense retriever using our constructed contrastive learning loss. This loss 
enables the retriever to effectively discriminate among positive, semi-positive, and negative samples, which are categorized based on their relevance to the query and the overshadowed keyphrase. As a result, the retriever achieves improved retrieval of documents relevant to the overshadowed keyphrase while avoiding query-irrelevant noise. 
({\romannumeral3}) \textit{Query Formulation}: Given the retrieved documents, we prompt the LLM to select the most relevant one and generate a new query that articulates the next reasoning step. 
In summary, ActiShade supplements the overshadowed knowledge when generating the query for the next-round retrieval, while minimizing the introduction of query-irrelevant noise, thereby reducing the error accumulation caused by \textit{knowledge overshadowing}.

We evaluate ActiShade on three widely used multi-hop reasoning datasets: HotpotQA~\cite{hotpotqa}, 2WikiMQA~\cite{2wiki}, and MuSiQue~\cite{musique}. 
Experimental results show that ActiShade significantly outperforms the state-of-the-art baselines on three datasets. 
Our contributions can be summarized as follows:
\begin{itemize} 
    \item We propose ActiShade, a novel framework designed to detect and leverage overshadowed knowledge for multi-hop reasoning. 
    \item In ActiShade, we design a new Gaussian perturbation-based method, GaP, to detect the overshadowed knowledge. 
    \item In ActiShade, we introduce a novel contrastive learning loss for retriever training and a query formulation strategy to leverage the overshadowed knowledge.  
    \item We conduct comprehensive experiments on three datasets, demonstrating that ActiShade outperforms the state-of-the-art methods across multiple LLMs in terms of effectiveness. 
\end{itemize}

\section{Related Work}

\subsection{Knowledge Overshadowing}
\citet{zhang2024knowledge} observed when extracting knowledge from LLMs using queries involving multiple conditions, some conditions may overshadow others, causing them to be ignored and thus leading to hallucinated outputs-a phenomenon they refer to as \textit{knowledge overshadowing}. 
This phenomenon is present in multi-hop QA scenarios, which limits the effectiveness of multi-round retrieval approaches. 
Specifically, \textit{knowledge overshadowing} causes LLMs to generate factually incorrect outputs. As multi-round retrieval methods typically rely on LLM-generated output as the next-round retrieval query, such hallucinations lead to irrelevant retrieval and cause error accumulation during the iterative
process. 
To overcome this limitation, we propose a novel multi-round retrieval framework, ActiShade, which reduces the error accumulation caused by \textit{knowledge overshadowing}. 
\citet{zhang2024knowledge} also proposed CoDA to detect overshadowed knowledge by removing tokens from the query and measuring changes in the output distribution. In contrast, our GaP preserves the reasoning chain by adding Gaussian noise instead of removing tokens, which enhances the detection of \textit{knowledge overshadowing} in multi-hop reasoning. 

\begin{figure*}[ht]
    \centering
        \includegraphics[height=0.43\textheight]{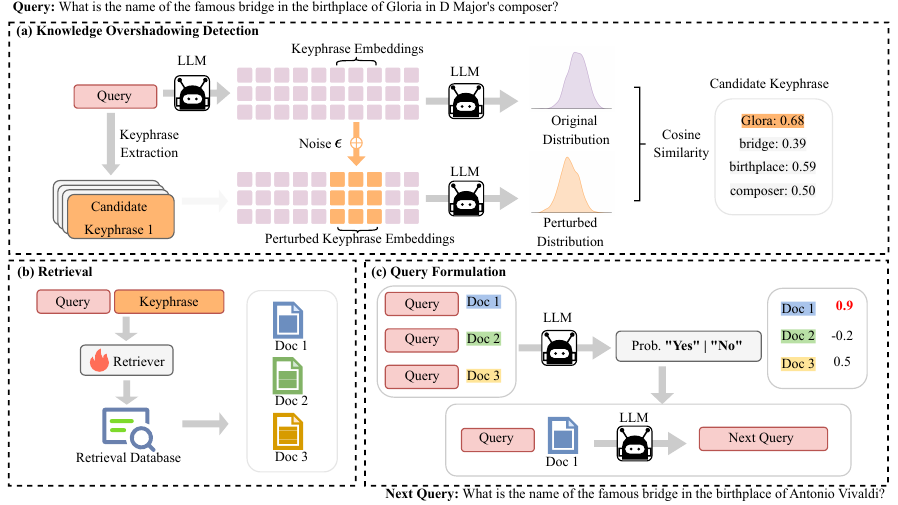} 
        \caption{Overview of ActiShade. ActiShade first detects the overshadowed keyphrase in the query, then retrieves relevant documents based on it, and finally formulates a new query for the next-round retrieval.
        } 
        \label{fig:framework}
\end{figure*}

\subsection{Retrieval-augmented LLM}
LLMs have a risk of generating hallucinated responses, thus necessitating external retrieval for retrieval-augmented generation. 
Previous methods typically adopt one-round retrieval, \textit{i.e.}, retrieve knowledge using only the original question once~\cite{DBLP:conf/icml/GuuLTPC20, DBLP:conf/icml/BorgeaudMHCRM0L22, DBLP:conf/emnlp/ZhangZRSHWLC23,DBLP:journals/jmlr/IzacardLLHPSDJRG23,DBLP:conf/naacl/ShiMYS0LZY24}. 
This line of work, however, struggles to gather all the necessary knowledge to answer multi-hop questions. 
Recently, another line of work arose, which adopts multi-round retrieval to meet multi-hop knowledge needs. 
SelfASK~\cite{self-ask} prompts LLMs to decompose a complex question into sub-questions and answer them through a search engine. 
Iter-RetGen~\cite{iter-retgen} leverages the output from the previous round concatenated with the question as the query for next-round retrieval. 
IRCoT~\cite{ircot} uses each CoT sentence as a query for retrieval until the final answer is obtained. 
FLARE~\cite{flare} determine when to retrieve based on reasoning confidence. 
BeamAggR~\cite{DBLP:conf/acl/ChuCCWZDYLQ24} decomposes complex questions, then performs bottom-up multi-source reasoning via post-order traversal, and uses beam aggregation to obtain the final answer. As it relies on multi-source knowledge, which differs from our setting, we do not include it as a baseline for fair comparison. 
EfficientRAG~\cite{zhuang-etal-2024-efficientrag} iteratively generates new retrieval queries and filters out irrelevant information using small models. 
DRAGIN~\cite{su-etal-2024-dragin} decides when and what to retrieve based on the LLM’s information needs during the generation process. 

Compared to them, our method is designed to reduce the error accumulation caused by \textit{knowledge overshadowing} and shows superior performance. Besides, we propose a novel method to detect overshadowed keyphrases through noise perturbation. 

\section{ActiShade Framework}

In this section, we introduce ActiShade, a novel multi-round retrieval framework that aims to reduce error accumulation caused by \textit{knowledge overshadowing}. 
ActiShade consists of three modules: (1) \textit{Knowledge Overshadowing Detection} for detecting the overshadowed keyphrase; (2) \textit{Retrieval based on Overshadowed Keyphrase} for relevant document retrieval; and (3) \textit{Query Formulation} for next-round retrieval query generation. 
An overview of the framework is illustrated in Figure~\ref{fig:framework}. 

\subsection{Knowledge Overshadowing Detection}
In this module, we propose a novel method, GaP, to detect \textit{knowledge overshadowing} in the query. The method consists of three steps: keyphrase extraction, keyphrase perturbation, and knowledge overshadowing measuring. 

\paragraph{Step 1. Keyphrase Extraction.} 
Given a query $Q$, we extract a set of candidate keyphrases $P=\{p_1, p_2, ..., p_n\}$, where each $p_i$ is a span within $Q$. Specifically, we utilize SpaCy~\cite{honnibal2017spacy} to extract named entities and meaningful tokens with POS tags in the set $\{\texttt{NOUN, ADJ, VERB, PROPN, NUM, ADV}\}$, and remove stopwords to reduce noise.

\paragraph{Step 2. Keyphrase Perturbation.} 
For each keyphrase $p_i \in P$, we inject Gaussian noise into its token embeddings while keeping other tokens unchanged. The perturbed input embeddings are computed as:
\begin{equation}
\tilde{\textbf{H}}_{p_i} = \textbf{H} + \bm{m}_{p_i} \odot \bm{\epsilon}, \quad \bm{\epsilon} \sim \mathcal{N}(0, \sigma^2)
\label{eq:perturbed_embedding},
\end{equation}
where $\textbf{H}$ denotes the original input embeddings of the query, $\bm{m}_{p_i}$ is a binary mask that takes the value 1 at token positions corresponding to the keyphrase $p_i$ and 0 elsewhere, and $\bm{\epsilon}$ is Gaussian noise with zero mean and standard deviation $\sigma$. 

We then input $\tilde{\textbf{H}}_{p_i}$ into the LLM to generate the perturbed output distribution: 
\begin{equation}
\tilde{\textbf{O}}_{p_i} = \mathbb{P}(y \mid \tilde{\textbf{H}}_{p_i}), \quad i = 1, 2, ..., n
\label{eq:perturbed_output}
\end{equation}

For comparison, the unperturbed output distribution is given by:
\begin{equation}
\textbf{O} = \mathbb{P}(y \mid \textbf{H}),
\label{eq:original_output}
\end{equation}
where $y$ denotes the LLM's output. 

\paragraph{Step 3. Knowledge Overshadowing Measuring.}  
To assess the influence of each keyphrase $p_i \in P$ on the LLM output, we first apply average pooling to the original and perturbed output distributions, $\textbf{O}$ and $\tilde{\textbf{O}}_{p_i}$, along the temporal dimension, resulting in pooled representations $\textbf{r}$ and $\tilde{\textbf{r}}_{p_i}$.

We then compute the cosine similarity between $\textbf{r}$ and $\tilde{\textbf{r}}_{p_i}$, and consider the keyphrase with the highest similarity to be overshadowed:
\begin{equation}
p_{ko} = \arg\max_{p_i \in P} \cos(\textbf{r}, \tilde{\textbf{r}}_{p_i})
\label{eq:argmax_cosine}
\end{equation}

A high similarity score indicates that the perturbation had minimal influence on the LLM's output, suggesting that the keyphrase is underutilized, i.e., overshadowed. 

Our method applies perturbations rather than removing tokens from the query, thereby preserving the structure of the query, which enhances the detection of knowledge overshadowing.

\subsection{Retrieval based on Overshadowed Keyphrase} 
Given the detected overshadowed keyphrase $p_{ko}$, this module retrieves the documents that are relevant to both the query and the overshadowed keyphrase. 
To enhance the retriever's ability to focus on the overshadowed keyphrase and avoid introducing query-irrelevant noise, we propose a novel contrastive learning loss and train a dense retriever to discriminate three types of documents: positive (relevant to both the query and the keyphrase), semi-positive (relevant to the query but not the keyphrase), and negative (irrelevant to both).

\paragraph{Data Preparation.} 
We construct our training dataset based on the MuSiQue~\cite{musique} benchmark. Each data in the MuSiQue dataset is formulated in a dictionary format with the keys \texttt{question\_decomposition}, \texttt{question}, and \texttt{paragraphs}. 
The \texttt{paragraphs} field contains a set of documents that are either relevant or irrelevant to the question.  
The \texttt{question\_decomposition} field provides a list of sub-questions derived from the original question, each annotated with the supporting document required to answer it, which can be found in the \texttt{paragraphs} set. 

We first identify the subject entity of the first sub-question and define it as the keyphrase. 
The supporting document associated with the first sub-question is labeled as the \textbf{positive document ($D^+$)}.
The supporting documents for other sub-questions are labeled as \textbf{semi-positive documents ($D^*$)}, as they are necessary for answering the original question but are not directly related to the keyphrase. All remaining documents are labeled as \textbf{negative document ($D^-$)}, which are irrelevant to the question. 
Due to space limitations, annotation examples can be found in the Appendix.


\paragraph{Loss Function Construction.} We extend the contrastive loss proposed by~\cite {izacard2021unsupervised} to improve the capability of the retriever to prioritize documents relevant to both the query and a specified phrase within it. 
The loss function $\mathcal{L}$ is defined as follow: 

\begin{equation}
\textstyle
\mathcal{L}_1 = -\log \frac{S(Q, D^+)}{S(Q, D^+) + \sum S(Q, D^*) + \sum S(Q, D^-)},
\end{equation}

\begin{equation}
\textstyle
\mathcal{L}_2 = -\log \frac{S(Q, D^+) + \sum S(Q, D^*)}{S(Q, D^+) + \sum S(Q, D^*) + \sum S(Q, D^-)},
\end{equation}

\begin{equation}
\textstyle
\mathcal{L} = \alpha\mathcal{L}_1 + (1-\alpha)\mathcal{L}_2
\end{equation}

For brevity, we denote $S(Q, D) = e^{\text{sim}(Q, D)}$, where $\text{sim}$ indicates the cosine similarity, and introduce hyperparameter $\alpha$ to balance the loss terms. 
The loss $\mathcal{L}_1$ encourages positive pairs to have higher scores over both semi-positive and negative pairs. 
Although semi-positive documents $D^*$ are not directly relevant to any phrase in the question, they are required to answer the question and thus are more important than negative documents $D^-$. 
We introduce loss $\mathcal{L}_2$ to further distinguish semi-positive documents from negative ones. 
The combined loss $\mathcal{L}$ ensures the retriever ranks documents in the desired order: $ D^+> D^*> D^-$.

\paragraph{Retrieval.} We concatenate the query $Q$ with its corresponding overshadowed keyphrase $p_{ko}$ as input to retrieve a set of relevant documents $RD=\{rd_1, rd_2, ..., rd_n\}$. 
The trained retriever is capable of retrieving documents relevant to both the query and the overshadowed keyphrase, ensuring that the retrieved documents are not only query-relevant but also supplement the overshadowed knowledge, thereby enhancing LLMs' reasoning. 

\subsection{Query Formulation}
The previous module returns a set of retrieved documents $RD$, which is relevant to both the query and the overshadowed keyphrase within it. 
This module then formulate a new query based on the retrieved documents for the subsequent retrieval round. 
The query formulation process consists of three steps: relevant document selection, query generation, and subsequent-round retrieval decision. 

\paragraph{Step 1. Relevant Document Selection.} 
Given a collection of retrieved documents $RD$, we first prompt the LLM to select the most relevant one $rd_{m}$. 
Specifically, each retrieved document and the query are jointly input into the LLM. The LLM is required to determine whether the retrieved document is relevant to the query. If it is relevant, output ``Yes''; otherwise, output ``No''. 
A higher probability assigned to ``Yes'' suggests a higher degree of relevance. 
The most relevant retrieved document is then selected based on the probability of outputting ``Yes''. 
The prompt template used for this step is detailed in the Appendix. 

\begin{table*}[ht]
\centering
\small
\begin{tabular}{clccccccccccc}
\toprule
\multirow{2}{*}{\textbf{Model}} & \multirow{2}{*}{\textbf{Method}} 
& \multicolumn{3}{c}{\textbf{MuSiQue}} & & \multicolumn{3}{c}{\textbf{HotpotQA}} & & \multicolumn{3}{c}{\textbf{2WikiMQA}} \\
\cmidrule{3-5} \cmidrule{7-9} \cmidrule{11-13}
~ & ~ & ACC & & F1 & & ACC & & F1 & & ACC & & F1 \\
\hline
\multirow{9}{*}{\textbf{Llama-3-8B-Instruct}}
& \textbf{Direct} & 5.60 & & 9.96  &  & 22.40 & & 25.34  &  & 26.60 & & 31.25   \\
& \textbf{CoT} & 11.65 & & 16.29  &  & 29.00 & & 34.09  &  & 27.60 & & 34.19  \\
& \textbf{Direct-R}\textsuperscript{$\heartsuit$} & 11.42 & & 16.06 &  & 37.7 & & 44.89  &  & 28.37 & & 35.56   \\
& \textbf{Iter-RetGen}\textsuperscript{$\clubsuit$} & 18.24 & & 20.59 &  & 48.23 & & 49.41  &  & 38.71 & & 44.56   \\
&  \textbf{IRCoT}\textsuperscript{$\clubsuit$} & 15.57 & & 18.32 &  & 40.10 & & 47.03  &  & 34.20 & & 41.01   \\
& \textbf{SelfASK}\textsuperscript{$\clubsuit$} & 20.60 & & 21.41 & & 47.10 & & 48.70 & & 39.50 & & 43.87   \\
& \textbf{FLARE}\textsuperscript{$\clubsuit$} & 19.74 & & 20.50  & & 48.45 & & 50.40  & & 41.35 & & 42.24   \\
& \textbf{DRAGIN}\textsuperscript{$\clubsuit$} & 21.11 & & 22.61 &  & 50.87 & & 52.52  & & 40.78 & & 42.31   \\
& \textbf{ActiShade (Ours)} & \textbf{25.25} & & \textbf{26.94} & &  \textbf{54.60} & & \textbf{56.33}  & & \textbf{45.80} & & \textbf{46.02}   \\
\hline
\multirow{9}{*}{\textbf{Qwen2.5-7B-Instruct}}
& \textbf{Direct} & 3.80 & & 11.09  &  & 19.40 & & 19.52  &  & 26.80 & & 29.95   \\
& \textbf{CoT} & 6.00 & & 13.93  &  & 22.00 & & 27.61  &  & 29.00 & & 32.24  \\
& \textbf{Direct-R}\textsuperscript{$\heartsuit$} & 11.60 & & 17.24 &  & 43.00 & & 47.99  &  & 38.60 & & 41.17   \\
& \textbf{Iter-RetGen}\textsuperscript{$\clubsuit$} & 15.40 & & 18.07 &  & 44.40 & & 48.24  &  & 41.20 & & 42.98   \\
& \textbf{IRCoT}\textsuperscript{$\clubsuit$} & 14.90 & & 18.01 &  & 43.79 & & 48.13  &  & 40.21 & & 40.84   \\
& \textbf{SelfASK}\textsuperscript{$\clubsuit$} & 17.35 & & 20.60 & & 42.18 & & 46.10 & & 43.17 & & 44.30   \\
& \textbf{FLARE}\textsuperscript{$\clubsuit$} & 10.69 & & 14.89  & & 40.19 & & 42.03  & & 39.21 & & 40.80  \\
& \textbf{DRAGIN}\textsuperscript{$\clubsuit$} & 19.80 & & 22.01 &  & 46.10 & & 50.30  & & 45.90 & & 45.87   \\
& \textbf{ActiShade (Ours)} & \textbf{22.80} & & \textbf{26.11} & &  \textbf{48.20} & & \textbf{55.45}  & & \textbf{52.80} & & \textbf{50.47}   \\
\hline
\multirow{9}{*}{\textbf{Qwen2.5-14B-Instruct}}
& \textbf{Direct} & 6.20 & & 13.48  &  & 27.20 & & 32.94  &  & 29.00 & & 32.39   \\
& \textbf{CoT} & 10.40 & & 17.74  &  & 29.40 & & 34.92  &  & 33.40 & & 35.76  \\
& \textbf{Direct-R}\textsuperscript{$\heartsuit$} & 14.80 & & 19.16 &  & 46.20 & & 48.11  &  & 39.00 & & 43.14  \\
& \textbf{Iter-RetGen}\textsuperscript{$\clubsuit$} & 17.20 & & 21.54 &  & 49.03 & & 51.04 &  & 43.20 & & 45.19   \\
& \textbf{IRCoT}\textsuperscript{$\clubsuit$} & 15.89 & & 19.90 &  & 47.98 & & 50.50 &  & 44.60 & & 45.71   \\
& \textbf{SelfASK}\textsuperscript{$\clubsuit$} & 18.48 & & 21.75 & & 45.97 & & 47.19 & & 47.49 & & 49.13   \\
& \textbf{FLARE}\textsuperscript{$\clubsuit$} & 13.10 & & 18.00 & & 42.14 & & 44.87  & & 40.01 & & 40.74  \\
& \textbf{DRAGIN}\textsuperscript{$\clubsuit$} & 22.70 & & 24.11 &  & 51.21 & & 54.30 & & 48.10 & & 49.87  \\
& \textbf{ActiShade (Ours)} & \textbf{25.59} & & \textbf{27.47} & &  \textbf{53.97} & & \textbf{57.45}  & & \textbf{51.13} & & \textbf{53.29}   \\
\bottomrule
\end{tabular}
\caption{The overall experimental results of ActiShade and other baselines on three benchmarks. The best results are in bold. $\heartsuit$ donates single-round retrieval. $\clubsuit$ indicates multi-round retrieval. }
\label{main_result}
\end{table*}

\paragraph{Step 2. Query Generation.} 
In the second step, we prompt the LLM to generate a new query $Q_{next}$ based on the most relevant retrieved document $rd_m$. 
The newly generated query is used for the subsequent retrieval round, aiming to retrieve more information beyond the scope of the initial query. The prompt template used for this step is detailed in Appendix. 
Figure~\ref{fig:framework} presents examples of query generation. 

Since the retrieved document serves to supplement the overshadowed knowledge, it enable the generation of a more accurate query. Moreover, the newly generated query explicitly presents implicit reasoning results. 
These allow the new query to lead to more accurate and relevant retrieval in the next round, thereby reducing the error accumulation caused by \textit{knowledge overshadowing}. 

\paragraph{Step 3. Subsequent-Round Retrieval Decision.} 
To decide whether to terminate the iterative process, we prompt the LLM to assess whether more information is needed to answer the initial query.
Specifically, the new query $Q_{next}$ is input into the LLM, which is required to determine whether it is a single-hop query. If it is, we perform an additional retrieval round and then terminate; otherwise, the retrieval continues iteratively.  
The iterative process also terminates if the maximum number of iterations is reached. 
The prompt template used for this step is detailed in the Appendix. 

After iteration, we input the initial query and the relevant documents retrieved during the iterative process into the LLM to obtain the final response.

\section{Experimental Setup}
\subsection{Datasets}
We evaluate ActiShade on three multi-hop reasoning datasets. HotpotQA~\cite{hotpotqa}, 2WikiMQA~\cite{2wiki} consist of two-hop questions, and MusiQue~\cite{musique} contains questions with 2 to 4 hops. For HotpotQA, 2WikiMQA, and MuSiQue,  we use the same test set provided by IRCoT~\cite{ircot}, which contains 500 randomly sampled instances from the original development set. 

\subsection{Implementation Details}
We use Llama-3-8B-Instruct~\cite{llama3} and Qwen2.5-Instruct (7B and 14B)~\cite{qwen} as the backbone language models for generation and reasoning.
For retriever training, we fine-tune \texttt{contriever-msmarco}~\cite{izacard2021unsupervised} on a subset of the MuSiQue training set. We manually select 5{,}000 high-quality examples, of which 3{,}500 are used for training, 750 for validation, and 750 for testing. The retriever is trained using the AdamW optimizer (learning rate 5e-5, batch size 32) for up to 20 epochs with early stopping based on validation loss. The contrastive loss combines two objectives with a weighting coefficient $ \alpha = 0.7 $. 
All experiments are conducted on two NVIDIA A6000 GPUs. 

\subsection{Baselines}
We choose the following methods as baselines. \textbf{Standard Prompting~\cite{sp}} directly generates the final answer.  \textbf{CoT Prompting~\cite{cot}} generates reasoning steps before the final answer.  \textbf{One-time Retrieval} retrieves relevant documents from external resources and incorporates them to generate the final answer.  \textbf{IRCoT~\cite{ircot}} alternates between retrieval-augmented reasoning and reasoning-augmented retrieval until enough information is obtained to answer the given question.  \textbf{Iter-RetGen~\cite{iter-retgen}} synergizes retrieval and generation in an iterative manner: the LLM's response serves as a query for retrieving more relevant knowledge, which in turn helps generate a better response in another iteration. \textbf{Self-Ask~\cite{self-ask}} iteratively decomposes complex questions into sub-questions, retrieves and answers them to reach the final answer.  \textbf{FLARE~\cite{flare}} dynamically adjusts retrieval timing based on reasoning confidence and retrieves guided by the upcoming reasoning sentences.  \textbf{DRAGIN~\cite{su-etal-2024-dragin}} detects information needs in real time and uses self-attention over context to form retrieval queries during the generation process.

\subsection{Evaluation Metrics}
For evaluation metrics, we utilize Accuracy (Acc) and F1 score metrics for evaluation. The ACC checks if the ground-truth answer is in the LLM-generated answer, which is also named Cover Exact Match. The F1 score is used to measure the overlap between the LLM-generated answer and the ground truth answer. 

\begin{table}[t]
\centering
\footnotesize
\setlength{\tabcolsep}{2pt} 
\begin{tabular}{p{2.6cm}ccc} 
\toprule
\textbf{Model / Dataset} & MuSiQue & HotpotQA & 2WikiMQA \\
\midrule
Direct-R & 16.06 & 44.89 & 35.56 \\
\ \ -\textit{w} CoDA & 15.86 & 45.98 & 35.49 \\
\ \ -\textit{w} GaP & \textbf{17.83} & \textbf{46.81} & \textbf{38.67} \\
\midrule
ActiShade-NoKOD & 22.83 & 51.23 & 45.18 \\
\ \ -\textit{w} CoDA & 21.23 & 52.45 & 41.29 \\
\ \ -\textit{w} GaP & \textbf{26.94} & \textbf{56.33} & \textbf{46.02} \\
\bottomrule
\end{tabular}
\caption{Performance comparison between GaP and CoDA in single-round and multi-round retrieval settings. ActiShade-NoKOD denotes ActiShade without the Knowledge Overshadowing Detection module. All results are reported in F1 score.
}
\label{Knowledge}
\end{table}

\section{Experimental Results}
\subsection{Main Results}
The experimental results on three multi-hop reasoning datasets are presented in Table~\ref{main_result}. We can obtain the following observations:
\subsubsection{Achieving Significant Performance Improvement across all datasets and LLMs.} 
ActiShade outperforms the previous state-of-the-art, DRAGIN, across all datasets and LLMs, highlighting its effectiveness in multi-hop reasoning. This performance improvement can be attributed to ActiShade's ability to reduce error accumulation caused by \textit{knowledge overshadowing} by iteratively detecting overshadowed keyphrases in the query, retrieving documents relevant to both the query and the overshadowed keyphrase, and generating a new query based on the retrieved documents. 
Notably, ActiShade surpasses SelfASK~\cite{self-ask}, which decomposes a complex question into sub-questions and answers them via retrieval. We believe this suggests that our query formulation process makes implicit reasoning explicit, enabling more accurate and relevant retrieval compared to question decomposition. 
\subsubsection{Maintaining Generalization Ability.} 
We train our retriever based on the MuSiQue dataset, as detailed in the ActiShade Framework section. 
However, ActiShade, on HotpotQA and 2WikiMQA, still outperforms all baselines across all LLMs, further demonstrating its effectiveness and generalization. 
This indicates that the retriever effectively learns to align retrieval not only with the query but also with the overshadowed keyphrase, allowing it to generalize well across various multi-hop reasoning benchmarks. 
\subsubsection{Effectiveness for larger models.}
To evaluate how effective ActiShade is at different model sizes, we conduct experiments on Qwen2.5-Instruct (7B and 14B). 
As shown in Table~\ref{main_result}, the ActiShade's performance generally improves with the model size, demonstrating its scalability to larger models. 
Due to hardware resource constraints, we are unable to implement ActiShade on larger models.

\begin{figure}[t]
    \centering
        \includegraphics[width=0.47\textwidth]{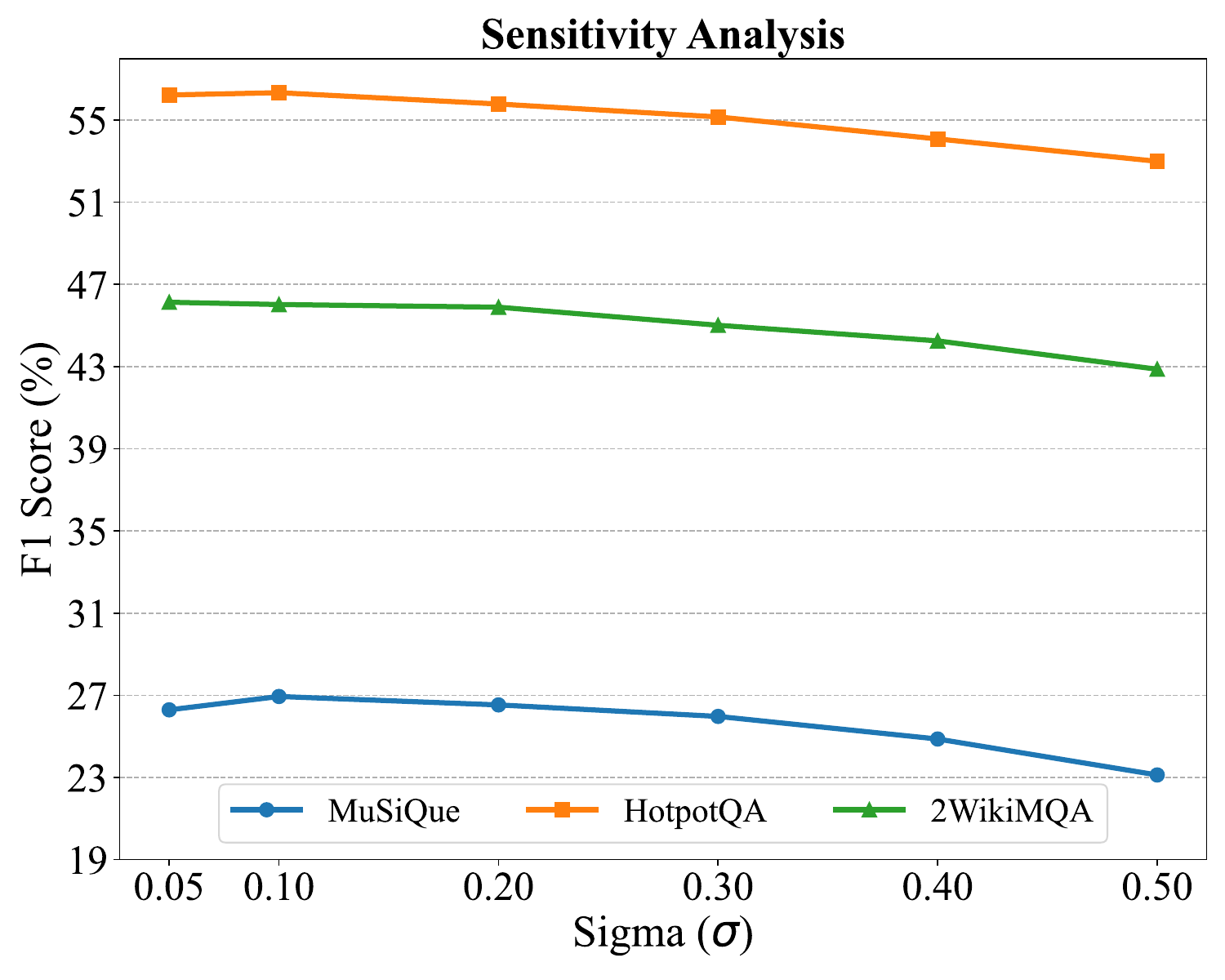} 
        \caption{Sensitivity analysis of the Gaussian noise standard deviation $\sigma$.} 
        \label{fig:sensitivity}
\end{figure}

\subsection{Analysis of Knowledge Overshadowing Detection}
To systematically evaluate the effectiveness of GaP, we conduct a series of analyses focusing on performance comparison, interpretability, and parameter sensitivity. 
\paragraph{Comparative Performance of Detection Methods.}
To investigate the impact of the knowledge overshadowing detection module, we compare our proposed GaP against the CoDA~\cite{zhang2024knowledge} method under both single-round and multi-round retrieval settings. 
In the single-round setting, we evaluate three variants: (1) Direct retrieval without overshadowing detection; (2) Direct retrieval with the CoDA integrated; (3) Direct retrieval with our GaP integrated. 
In the multi-round setting, we compare three corresponding setups: (1) ActiShade without the GaP method; (3) ActiShade replacing GaP with the CoDA method; (3) ActiShade (the full pipeline). 
The experimental results are shown in Table~\ref{Knowledge}. We observe that, in both single-round and multi-round retrieval settings, models incorporating our GaP method consistently outperform those without such integration, highlighting the effectiveness of GaP. 
In addition, we observe that, on the MuSiQue and 2WikiMQA datasets, models using CoDA even perform worse than those without knowledge overshadowing detection. 
This indicates that CoDA's token-removing approach may disrupt the reasoning chain in multi-hop questions, thereby limiting its effectiveness. 

\paragraph{Sensitivity and Interpretability Analysis of GaP.}
We conduct a sensitivity analysis to investigate how varying the standard deviation $\sigma$ of the Gaussian noise used for keyphrase perturbation affects the performance of our proposed ActiShade. 
All experiments in this analysis are conducted using the Llama-3-8B-Instruct, and the results are evaluated based on the F1 metric.  
As shown in Figure~\ref{fig:sensitivity}, we vary $\sigma$ in the range of [0.05, 0.5] and observe its impact on the final performance. Experimental results show that as $\sigma$ decreases, the model performance first improves, reaching a peak at $\sigma = 0.1$, and then gradually declines.
This indicates that a moderate level of noise can effectively help detect overshadowed keyphrases, while excessive noise causes large output distribution shifts for all candidate keyphrases, reducing the effectiveness of detection. 
Nevertheless, the overall performance remains relatively stable across a wide range of $\sigma$ values, suggesting that our method exhibits low sensitivity to this hyperparameter. 

\begin{figure}[t]
    \centering
        \includegraphics[width=0.47\textwidth]{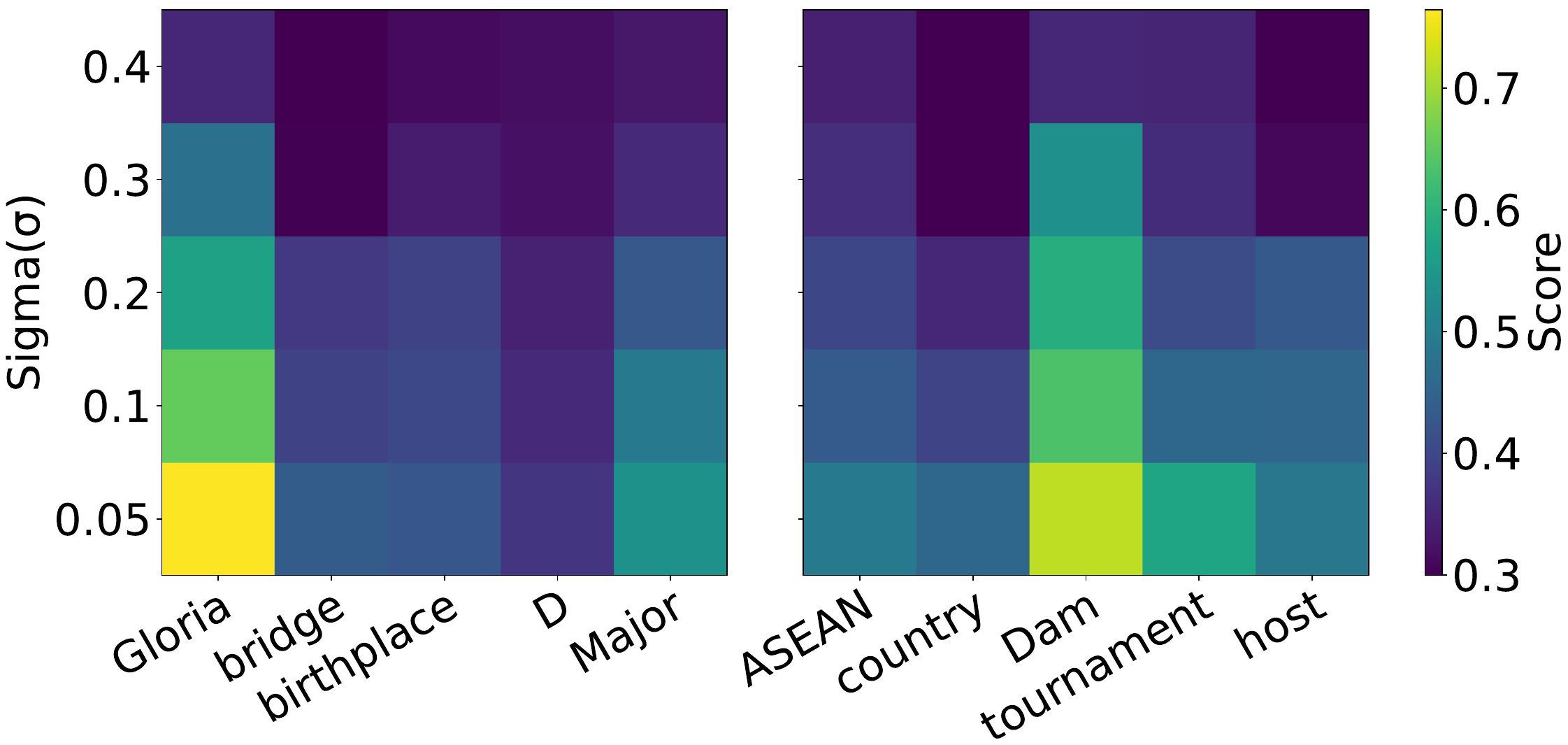} 
        \caption{Visualization analysis of the Gaussian noise standard deviation $\sigma$.} 
        \label{fig:heatmap}
\end{figure}

To interpret how GaP detects overshadowed keyphrases, we also conduct a visualization analysis of output distribution similarity across different keyphrases. We randomly select two queries and apply Gaussian noise of varying standard deviation to each candidate keyphrase. For each combination of keyphrase and noise level $\sigma$, we compute the similarity between the model's output distributions before and after perturbation. 
A high similarity suggests the keyphrase has little influence on the output and is likely overshadowed. Figure~\ref{fig:heatmap} shows that moderate perturbation best separates salient from overshadowed keyphrases, while stronger noise disrupts all outputs, lowers similarities across the board, and weakens detection.

\subsection{Analysis of Retriever Training}

We analyze the effectiveness of retriever training in the ActiShade framework by evaluating both the retrieval capability and the downstream QA performance under different training strategies. 

We first assess the retrieval ability of three retriever variants: (1) a retriever without task-specific fine-tuning (Base), (2) our proposed retriever trained with fine-grained contrastive learning (FCL), and (3) a retriever trained using standard contrastive learning (SCL) that only distinguishes between positive and negative examples. 
As shown in Table~\ref{tab:recall_grouped}, our method achieves the highest Recall@$k$ scores on both positive and semi-positive document retrieval, demonstrating its effectiveness in capturing multi-level document relevance critical for multi-hop reasoning. 
When distinguishing between positive\&semi-positive and negative examples, our method performs comparably to the retriever trained with standard contrastive learning on Recall@1, while outperforming it on Recall@3. 
This indicates that our improved contrastive learning objective helps the retriever better distinguish positive from semi-positive examples, while still effectively discriminating negative ones. 

We then examine how these retrievers affect the final QA performance. As presented in Table~\ref{retrieval}, our proposed retriever achieves the best results across three datasets. This shows that training the retriever to distinguish varying degrees of relevance is beneficial not only for retrieval capability but also for downstream answer generation. Notably, even without retriever training, ActiShade still outperforms previous baselines, highlighting the effectiveness of the Knowledge Overshadowing Detection and Query Formulation modules in the overall framework. 

\begin{table}[t]
\centering
\footnotesize
\begin{tabular}{lcccccc}
\toprule
\multirow{2}{*}{\textbf{Model}}
& \multicolumn{2}{c}{\textbf{Pos}} 
& \multicolumn{2}{c}{\textbf{Semi}} 
& \multicolumn{2}{c}{\textbf{Pos\&Semi}} \\
\cmidrule(lr){2-3} \cmidrule(lr){4-5} \cmidrule(lr){6-7}
& R@1 & R@3 & R@1 & R@3 & R@1 & R@3 \\
\midrule
Base  & 29.20 & 50.40 & 12.57 & 25.42 & 18.29 & 36.78 \\
SCL   & 57.84 & 69.21 & 40.12 & 59.99 & \textbf{38.21} & 50.29 \\
FCL  & \textbf{75.33} & \textbf{84.80} & \textbf{43.21} & \textbf{61.42} & 38.14 & \textbf{52.72} \\
\bottomrule
\end{tabular}
\caption{Comparison of Recall@1 and Recall@3 for different retrievers.}
\label{tab:recall_grouped}
\end{table}

\begin{table}[t]
\centering
\footnotesize
\setlength{\tabcolsep}{2pt} 
\begin{tabular}{p{2.7cm}ccc} 
\toprule
\textbf{Model / Dataset} & MuSiQue & HotpotQA & 2WikiMQA \\
\midrule
ActiShade & \textbf{26.94} & \textbf{56.33} & \textbf{46.02} \\
\ \ -\textit{w/} SCL & 24.10 & 54.25 & 44.97 \\
\ \ -\textit{w/o} FCL & 25.68 & 53.89 & 44.61 \\
\bottomrule
\end{tabular}
\caption{Evaluation of retriever training strategies in ActiShade. The performance is evaluated using the F1 score.}
\label{retrieval}
\end{table}

\section{Conclusion}
In this paper, we introduce ActiShade, a novel multi-round retrieval framework for multi-hop reasoning. 
ActiShade iteratively detects overshadowed keyphrases in the query, retrieves documents relevant to both the query and the overshadowed keyphrase, and generates a new query based on the retrieved documents for the next iteration, thereby reducing the error accumulation caused by \textit{knowledge overshadowing}. 
Extensive experiments demonstrate the effectiveness of ActiShade across multiple datasets and LLMs. 

\newpage

\section{Acknowledgements}
This work was supported by the National Key Research and Development Program of China (Grant No. 2024YFE0210800) and the National Natural Science Foundation of China (Grant No. 62476025).

\bibliography{aaai2026}

\newpage

\appendix

\section{Appendix for ActiShade}
\subsection{Details of Data Preparation} \label{dataexample}
We construct our training dataset for the retriever based on MuSiQue~\cite{musique}. 
Each data in MuSiQue dataset is formulated in a dictionary format with the key \texttt{question\_decomposition}, \texttt{question}, and \texttt{paragraphs}. 

The \texttt{question\_decomposition} field provides a list of sub-questions derived from the original question, each annotated with the supporting document required to answer it, which can be found in the \texttt{paragraphs} set. 
The \texttt{paragraphs} field contains a set of documents that are either relevant or irrelevant to the question.  
Documents that support any sub-question are considered relevant, while the rest are treated as irrelevant. 
An example is shown in the Table~\ref{tab:case_anno} to illustrate the data structure. 

For each data in MuSiQue, we first consider the subject entity of the first sub-question as the keyphrase; in the example, this is \textbf{Gloria in D Major}. 
The supporting document associated with the first sub-question is labeled as the \textbf{positive document ($D^+$)}—in this case, document \textbf{2}. 
The supporting documents for other sub-questions, which are necessary for answering the original question but not directly related to the keyphrase, are labeled as \textbf{semi-positive documents ($D^*$)}; here, documents \textbf{1} and \textbf{9} fall into this category. All remaining documents are labeled as \textbf{negative document ($D^-$)}, which are irrelevant to both the keyphrase and the original question. 

\begin{table*}[htbp] 
  \small
  \centering
  \begin{tabular}{lcc} 
    \toprule
    \textbf{question} & \multicolumn{2}{c}{What is the name of the famous bridge in the birthplace of \textbf{Gloria in D Major}'s composer?} \\ \midrule
    \multirow{7}{*}{\textbf{question\_decomposition}}
    & question & \textbf{Gloria in D Major} \textgreater\textgreater \  composer \\ 
    & paragraph\_support\_idx & \textbf{2} \\ \cmidrule{2-3} 
    & question & \#1 \textgreater\textgreater \  place of birth \\
    & paragraph\_support\_idx & \textbf{1} \\ \cmidrule{2-3} 
    & question & what is the name of the famous bridge in \#2 \\
    & paragraph\_support\_idx & \textbf{9} \\  \midrule
    \multirow{12}{*}{\textbf{paragraphs}}
    & idx & 0 \\
    & paragraph\_text & The Dufferin Street bridges are two... \\ \cmidrule{2-3} 
    & idx  & \textbf{1} \\
    & paragraph\_text & Orlando furioso RV 819... \\ \cmidrule{2-3} 
    & idx  & \textbf{2} \\
    & paragraph\_text & Antonio Vivaldi wrote at least three... \\ \cmidrule{2-3} 
    & \multicolumn{2}{c}{\textbf{...}} \\ \cmidrule{2-3} 
    & idx  & \textbf{9} \\
    & paragraph\_text & The Rialto Bridge... \\ \cmidrule{2-3} 
    & \multicolumn{2}{c}{\textbf{...}}  \\ \midrule
    \multicolumn{3}{c}{...} \\
    \bottomrule
    \end{tabular}
    \caption{A data example from the MuSiQue dataset.} 
    \label{tab:case_anno}
\end{table*}

\subsection{Details of LLMs}
The details of selected LLMs are as follows: 
\begin{itemize}
  \item Llama3 \cite{meta2024introducing} is a collection of pre-trained and instruction-tuned LLMs in 8 and 70B sizes. The instruction-tuned LLMs, called Llama-3-Instruction, are optimized for dialogue use cases. We selected \textbf{Llama-3-8B-Instruction}. 
  \item Qwen2.5 \cite{qwen} is a series of Qwen large language models, including both base and instruction-tuned versions, spanning a parameter scale from 0.5 billion to 72 billion. We selected \textbf{Qwen2.5-7B-Instruct} and \textbf{Qwen2.5-14B-Instruct}. 
\end{itemize} 

\subsection{Prompt Template}
The Query Formulation module of ActiShade formulate a new query based on the retrieved documentss for the next-round iteration. 
This process consists of three steps: relevant document selection, query generation, and subsequent-round decision. 
The prompt templates used in this process are detailed below. 
\subsubsection{Prompt Template for Relevant Document Selection}\label{prompt:step1}
\begin{figure*}[t]
    \centering
        \includegraphics[width=1\textwidth]
        {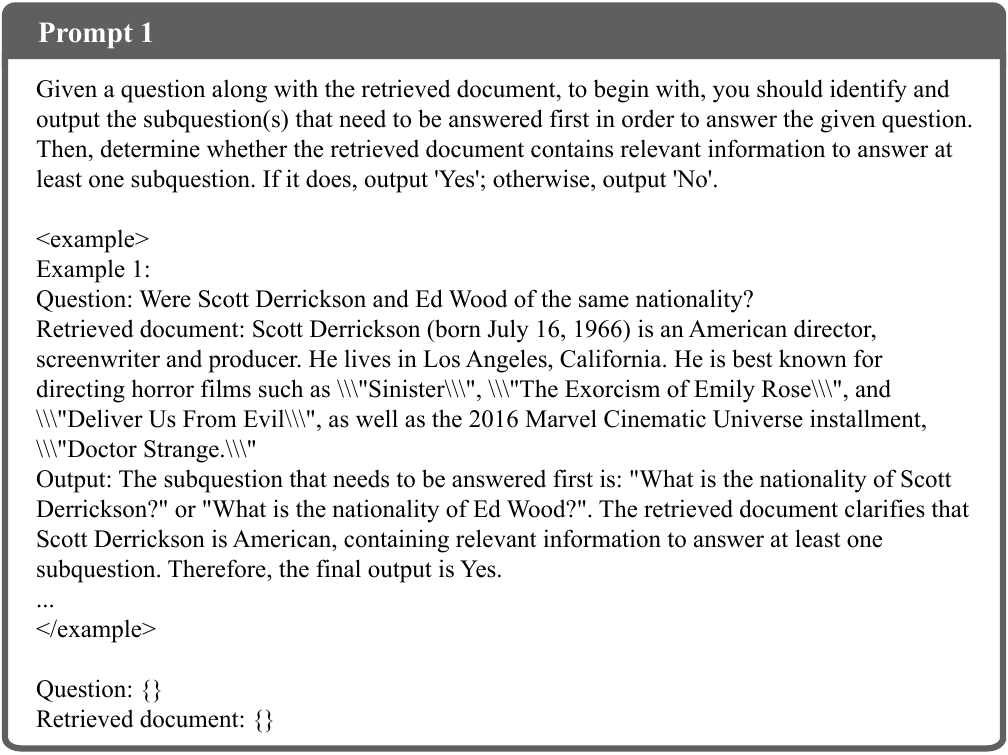} 
        \caption{Prompt template for relevant document selection in HotpotQA.} 
        \label{fig:hotpot_prompt1}
\end{figure*}

\begin{figure*}[t]
    \centering
        \includegraphics[width=1\textwidth]
        {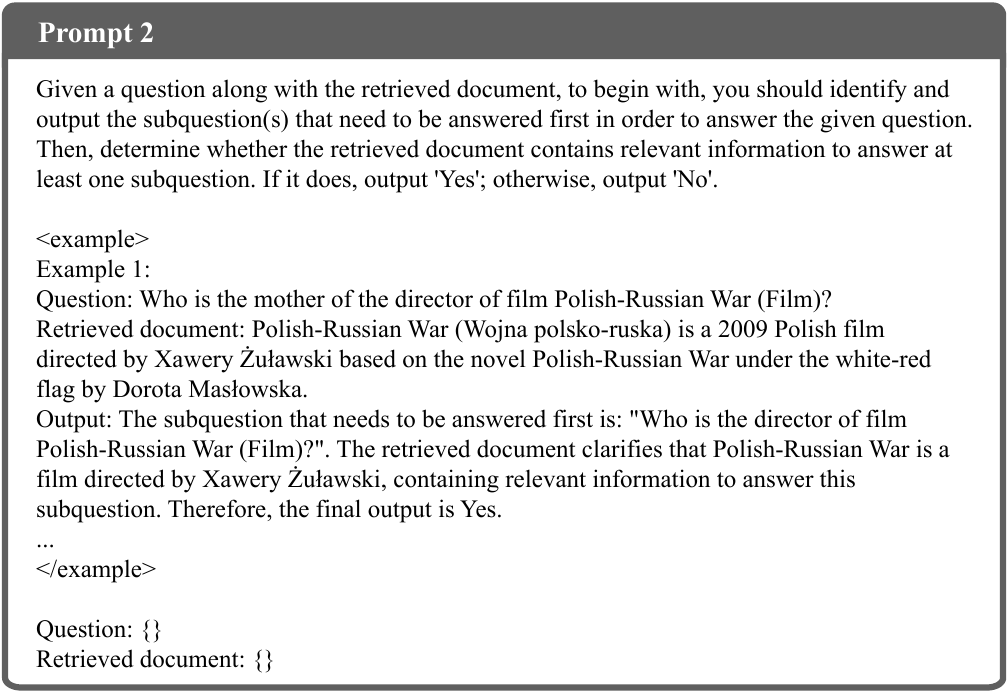} 
        \caption{Prompt template for relevant document selection in 2WikiQA.} 
        \label{fig:2wiki_prompt1}
\end{figure*}

\begin{figure*}[t]
    \centering
        \includegraphics[width=1\textwidth]
        {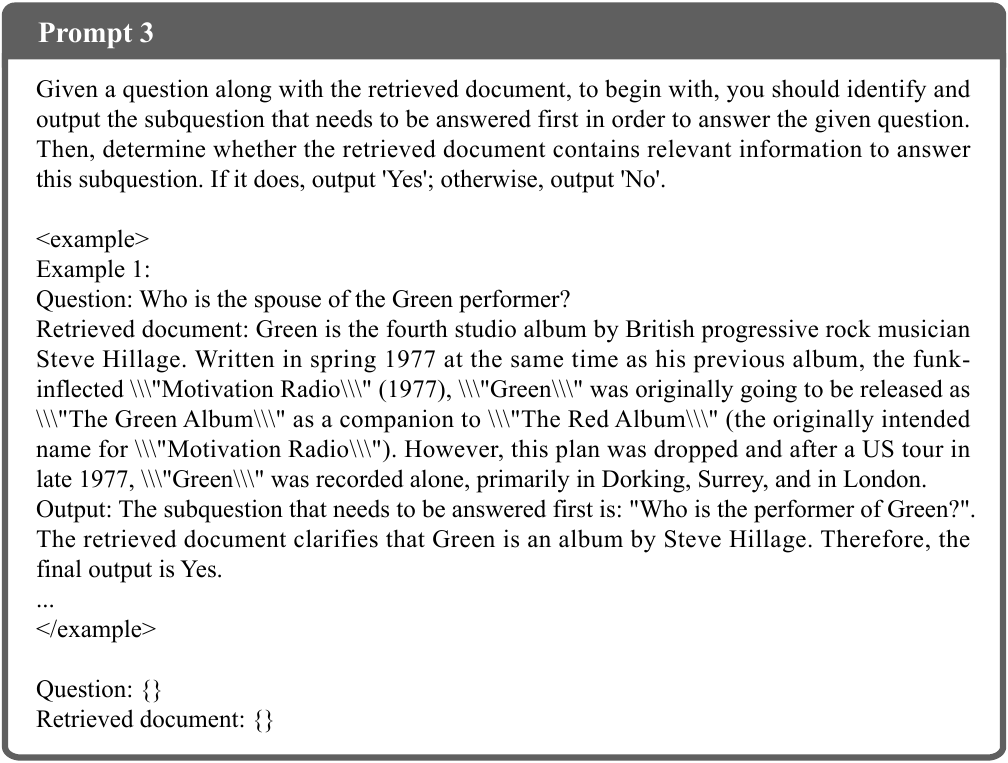} 
        \caption{Prompt template for relevant document selection in MuSiQue.} 
        \label{fig:musique_prompt1}
\end{figure*}

We first prompt the LLM to select the most relevant retrieved document through the prompt templates, as illustrated in Figure~\ref{fig:hotpot_prompt1}, Figure~\ref{fig:2wiki_prompt1}, and Figure~\ref{fig:musique_prompt1}. 

\subsubsection{Prompt Template for Query Generation}\label{prompt:step2}
\begin{figure*}[t]
    \centering
        \includegraphics[width=1\textwidth]
        {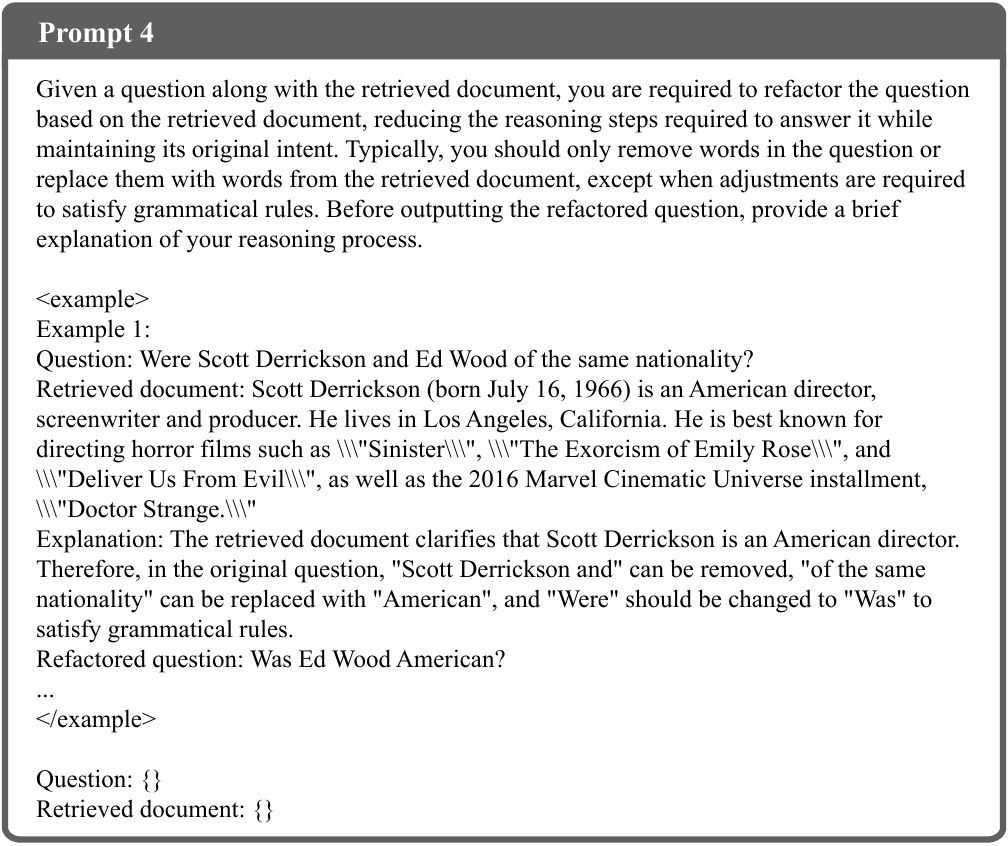} 
        \caption{Prompt template for query generation in HotpotQA.} 
        \label{fig:hotpot_prompt2}
\end{figure*}

\begin{figure*}[t]
    \centering
        \includegraphics[width=1\textwidth]
        {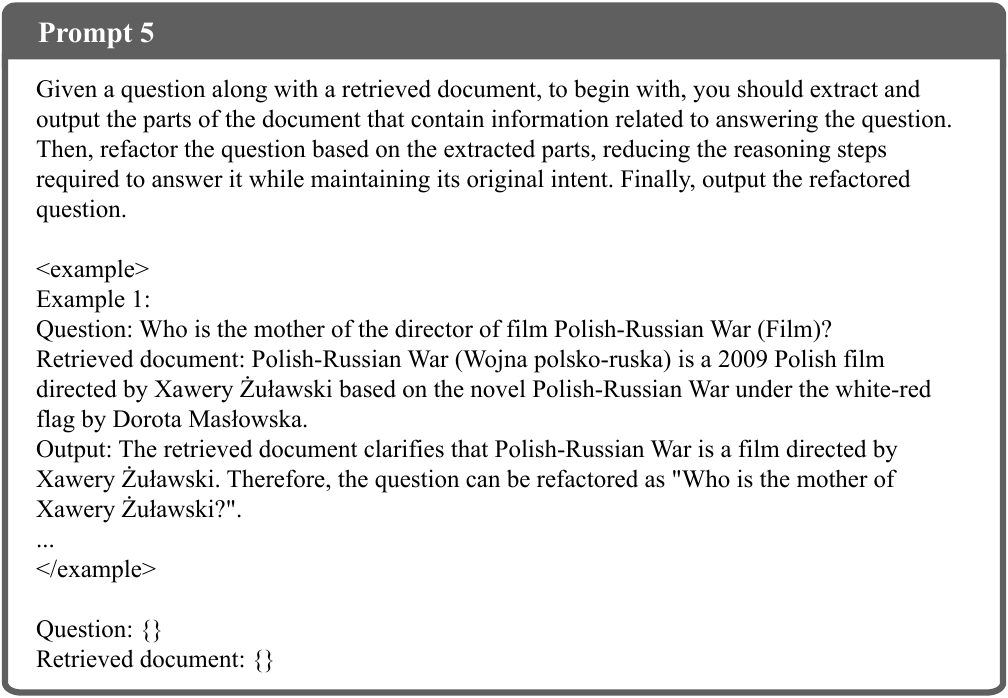} 
        \caption{Prompt template for query generation in 2WikiQA.} 
        \label{fig:2wiki_prompt2}
\end{figure*}

\begin{figure*}[t]
    \centering
        \includegraphics[width=1\textwidth]
        {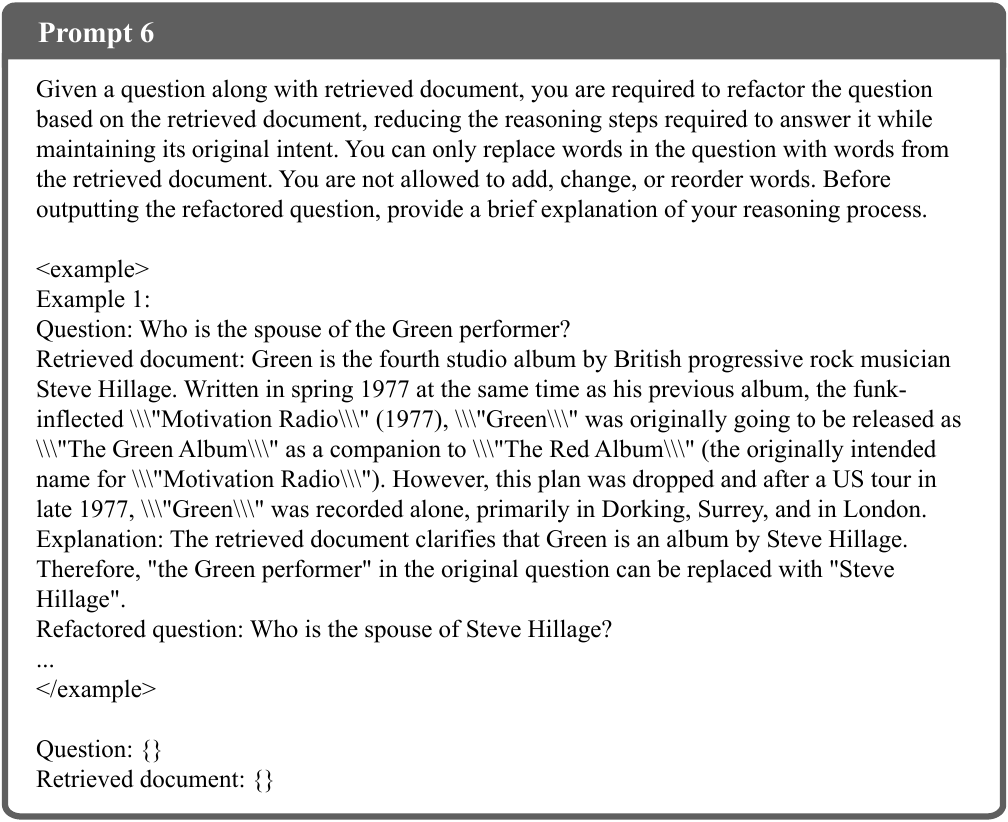} 
        \caption{Prompt template for query generation in MuSiQue.} 
        \label{fig:musique_prompt2}
\end{figure*}

We then prompt the LLM to generate a new query based on the most relevant retrieved document. This newly generated query explicitly presents implicit reasoning results and is used for the next-round iteration. The used prompt templates are shown in Figure~\ref{fig:hotpot_prompt2}, Figure~\ref{fig:2wiki_prompt2}, and Figure~\ref{fig:musique_prompt2}. 

\subsubsection{Prompt Template for Subsequent-Round Retrieval Decision}\label{prompt:step3}
\begin{figure*}[t]
    \centering
        \includegraphics[width=1\textwidth]
        {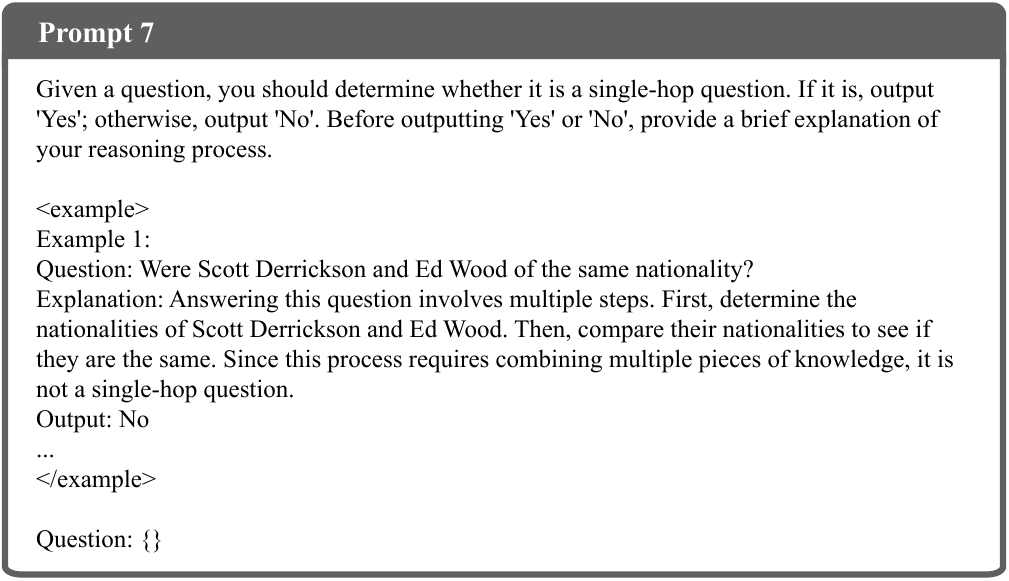} 
        \caption{Prompt template for subsequent-round retrieval decision in HotpotQA.} 
        \label{fig:hotpot_prompt3}
\end{figure*}

\begin{figure*}[t]
    \centering
        \includegraphics[width=1\textwidth]
        {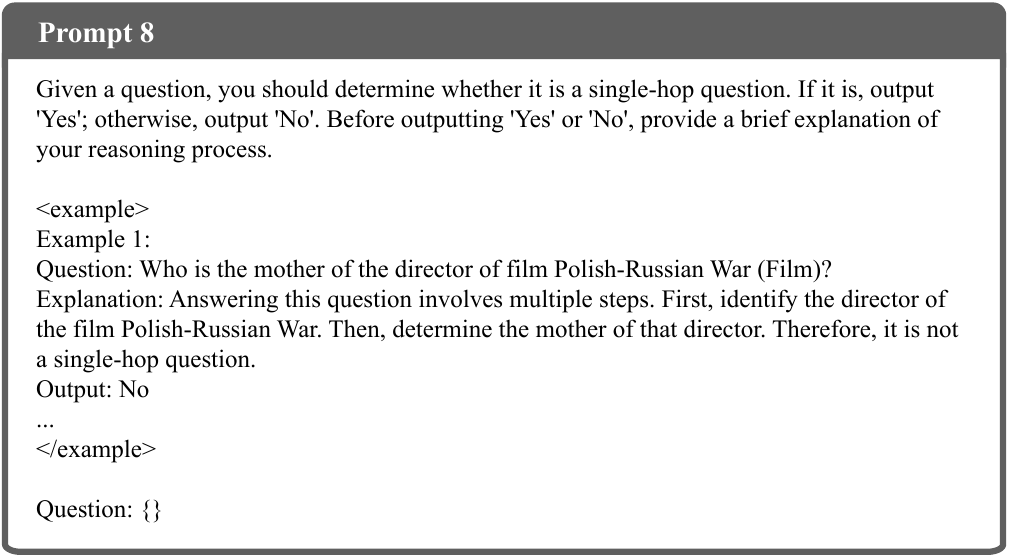} 
        \caption{Prompt template for subsequent-round retrieval decision in 2WikiQA.} 
        \label{fig:2wiki_prompt3}
\end{figure*}

\begin{figure*}[t]
    \centering
        \includegraphics[width=1\textwidth]
        {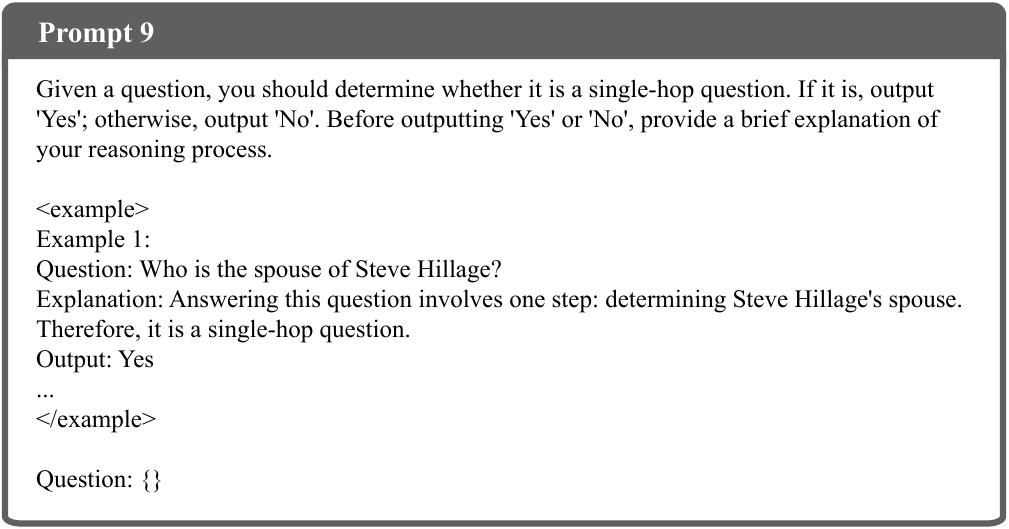} 
        \caption{Prompt template for subsequent-round retrieval decision in MuSiQue.} 
        \label{fig:musique_prompt3}
\end{figure*}

Finally, we prompt the LLM to determine whether the given query is single-hop. If it is, this indicates that sufficient information to answer the initial query can be obtained with one more round of retrieval. We then perform an additional retrieval round and terminate the process. The prompt templates used for this step are presented in Figure~\ref{fig:hotpot_prompt3}, Figure~\ref{fig:2wiki_prompt3}, and Figure~\ref{fig:musique_prompt3}. 

\subsection{Ablation Study on Query Formulation}

\begin{table}[t]
\centering
\footnotesize
\setlength{\tabcolsep}{2pt} 
\begin{tabular}{p{2.7cm}ccc} 
\toprule
\textbf{Model / Dataset} & MuSiQue & HotpotQA & 2WikiMQA \\
\midrule
ActiShade & \textbf{26.94} & \textbf{56.33} & \textbf{46.02} \\
\ \ -\textit{w/o} selection & 25.10 & 55.58 & 42.48 \\
\bottomrule
\end{tabular}
\caption{Ablation results on Query Formulation. The performance is evaluated using the F1 score.}
\label{retrieval}
\end{table}

We further examine the effectiveness of the relevant document selection step of the Query Formulation module. In this step, we prompt the LLM to select the most relevant document from the retrieved set to guide the generation of the next-round retrieval query. 
In the ablated variant, we directly use the document with the highest retrieval score, without performing relevant document selection. 
As shown in Table~\ref{retrieval}, excluding this step leads to a decline in performance. 
This suggests that the relevant document selection step ensures that the chosen document is beneficial for the following query generation step, thereby achieving better performance. 
\subsection{Case Study}
To illustrate how our method effectively mitigate the error accumulation caused by \textit{knowledge overshadowing} in multi-hop reasoning, we present a representative example as follows. 

\texttt{\textbf{Initial Question}: What is the name of the famous bridge in the birthplace of Gloria in D Major's composer?}

\noindent In the first round, the detected candidate keyphrases along with their corresponding scores are as follows: 

\texttt{Gloria 0.68, bridge 0.39, birthplace 0.59, composer 0.50.}

\noindent Based on these scores, we consider \texttt{Gloria} as the overshadowed keyphrase. After retrieval and relevant document selection, we obtain the following document: 

 \texttt{Title: Gloria (Vivaldi) — Antonio Vivaldi wrote at least three settings of the hymn Gloria in excelsis Deo ...}

\noindent The next-round query is then reformulated as: 

\texttt{What is the name of the famous bridge in the birthplace of Antonio Vivaldi?}

\noindent In this second iteration, the detected candidate keyphrases and their corresponding scores are listed as follows: 

\texttt{Antonio Vivaldi 0.63, bridge 0.15, birthplace 0.25}

\noindent Given these scores, we regard \texttt{Antonio Vivaldi} as the overshadowed keyphrase. After retrieval and relevant document selection, we obtain the following document: 

 \texttt{Title: Antonio Vivaldi — Antonio Lucio Vivaldi was an Italian Baroque musical composer. Born in Venice ...}

\noindent The third-round query is then reformulated as:

\texttt{What is the name of the famous bridge in Venice?}

\noindent Since this query is single-hop, we perform an additional round and then terminate iteration. The detected candidate keyphrases and scores include the following: 

\texttt{Venice: 0.79, bridge: 0.34;}

\noindent Given these scores, we consider \texttt{Venice} as the overshadowed keyphrase. After retrieval and relevant document selection, we obtain the following document: 

 \texttt{Title: Rialto Bridge — The Rialto Bridge is the oldest of the four bridges in Venice, Italy ...}

\noindent After iteration, we input the initial question along with the relevant documents retrieved during the iterative process into the LLM to generate the final answer \textbf{\texttt{Rialto Bridge}}.

\subsection{Comparison with Decomposition-Free Multi-Hop Retrieval Methods}
We also compare our method with decomposition-free multi-hop retrieval approaches. These methods aim to retrieve all supporting evidence in a single step by training the retriever to capture multi-hop relevance patterns. They primarily focus on enhancing the retriever’s ability to identify a set of documents that support complex questions. In contrast, our work is designed to improve the reasoning capability of large language models (LLMs). We adopt a retrieval-augmented generation (RAG) framework and introduce a multi-round interaction mechanism that enables the retriever and the LLM to collaborate iteratively. Specifically, our method detects keyphrases that might be overlooked by the LLM, retrieves documents relevant to these keyphrases, and reformulates the query to better guide subsequent reasoning. Therefore, our focus is on enhancing retrieval–generation synergy, rather than achieving one-shot retrieval accuracy. Moreover, these two lines of research are typically not directly compared in prior work, as they focus on different objectives: decomposition-free methods aim to improve retrieval accuracy and document coverage, while our method targets end-to-end QA performance by enhancing the interaction between retrieval and generation to support LLM reasoning.

Although our approach is conceptually different from decomposition-free methods, we include a comparison with representative baselines in this appendix to address the concerns raised by reviewers. We compare our method with several representative decomposition-free multi-hop retrieval approaches, including MDR~\cite{MDR}, Beam Retrieval~\cite{BeamRetrieval}, and GRITHopper~\cite{Grithopper}. For fair evaluation, all models are trained separately on the HotpotQA, 2WikiMQA, and MuSiQue datasets using their respective official implementations and training settings. We use the trained retrievers as plug-in retrieval modules and concatenate their top-retrieved documents with the original query, feeding the resulting input into the Llama-3-8B-Instruct model for answer generation. The experimental results are shown in  Table~\ref{free}. ActiShade consistently outperforms these decomposition-free baselines across all three datasets. 

\begin{table}[ht]
\centering
\footnotesize
\setlength{\tabcolsep}{2pt} 
\begin{tabular}{p{2.7cm}ccc} 
\toprule
\textbf{Model / Dataset} & MuSiQue & HotpotQA & 2WikiMQA \\
\midrule
MDR & 19.19 & 45.23 & 38.74 \\
Beam Retrieval & 19.79 & 47.37 & 38.98 \\
GritHopper & 22.13 & 50.76 & 41.30 \\
\midrule
ActiShade(ours) & \textbf{26.94} & \textbf{56.33} & \textbf{46.02} \\
\bottomrule
\end{tabular}
\caption{Comparison with decomposition-free multi-hop retrievers on three datasets. All results are reported in F1 score.}
\label{free}
\end{table}



\end{document}